\documentclass[sigconf]{acmart}

\usepackage{epsfig}
\usepackage{graphicx}
\usepackage{amsmath}

\usepackage{amssymb}
\usepackage{subfig}
\usepackage{makecell}
\usepackage{sidecap}

\usepackage[group-separator={,}]{siunitx}

\usepackage{balance}
\usepackage{bm}
\newcommand{\E}{\mathbb{E}}

\def \xx {{\bm{x}}}

\def \X  {\mathcal{X}}
\def \Y  {\mathcal{Y}}

\def \D  {\mathcal{D}}

\newcommand{\argmin}{\mathop{\mathrm{argmin}}}

\def\etal{\emph{et al. }}

\AtBeginDocument{%
  \providecommand\BibTeX{{%
    \normalfont B\kern-0.5em{\scshape i\kern-0.25em b}\kern-0.8em\TeX}}}

\copyrightyear{2020}
\acmYear{2020}
\setcopyright{acmcopyright}\acmConference[MM '20]{Proceedings of the 28th ACM International Conference on Multimedia}{October 12--16, 2020}{Seattle, WA, USA}
\acmBooktitle{Proceedings of the 28th ACM International Conference on Multimedia (MM '20), October 12--16, 2020, Seattle, WA, USA}
\acmPrice{15.00}
\acmDOI{10.1145/3394171.3413769}
\acmISBN{978-1-4503-7988-5/20/10}

\begin{document}
\fancyhead{}

\title{WildDeepfake: A Challenging Real-World Dataset for Deepfake Detection}

\author{Bojia Zi$^{1}$, Minghao Chang$^{1}$, Jingjing Chen$^{1}$, Xingjun Ma$^{2}$, Yu-Gang Jiang$^{1*}$}
\affiliation{%
  \institution{$^{1}$Shanghai Key Lab of Intelligent Information Processing, School of Computer Science, Fudan University}
  \institution{$^{2}$ School of Information Technology, Deakin University, Geelong, Australia}
  }
  
\thanks{$^*$ indicates corresponding author.}

\begin{abstract}
   In recent years, the abuse of a face swap technique called deepfake \cite{Deepfake} has raised enormous public concerns.
   So far, a large number of deepfake videos (known as "deepfakes") have been crafted and uploaded to the internet, calling for effective countermeasures.
   One promising countermeasure against deepfakes is deepfake detection.
   Several deepfake datasets have been released to support the training and testing of deepfake detectors, such as DeepfakeDetection \cite{DeepfakeDetection} and FaceForensics++ \cite{FaceForensics}.
   While this has greatly advanced deepfake detection, most of the real videos in these datasets are filmed with a few volunteer actors in limited scenes, and the fake videos are crafted by researchers using a few popular deepfake softwares. Detectors developed on these datasets may become less effective against real-world deepfakes on the internet.
   To better support detection against real-world deepfakes, in this paper, we introduce a new dataset \textbf{WildDeepfake}, which consists of \textbf{7,314} face sequences extracted from \textbf{707} deepfake videos collected completely from the internet. WildDeepfake is a small dataset that can be used, in addition to existing datasets, to develop and test the effectiveness of deepfake detectors against real-world deepfakes.
   We conduct a systematic evaluation of a set of baseline detection networks on both existing and our WildDeepfake datasets, and show that WildDeepfake is indeed a more challenging dataset, where the detection performance can decrease drastically. We also propose two (eg. 2D and 3D) Attention-based Deepfake Detection Networks (ADDNets) to leverage the attention masks on real/fake faces for improved detection. We empirically verify the effectiveness of ADDNets on both existing datasets and WildDeepfake. The dataset is available at: \url{https://github.com/OpenTAI/wild-deepfake}.
\end{abstract}


\begin{teaserfigure}
  \centering
  \includegraphics[width=0.72\textwidth]{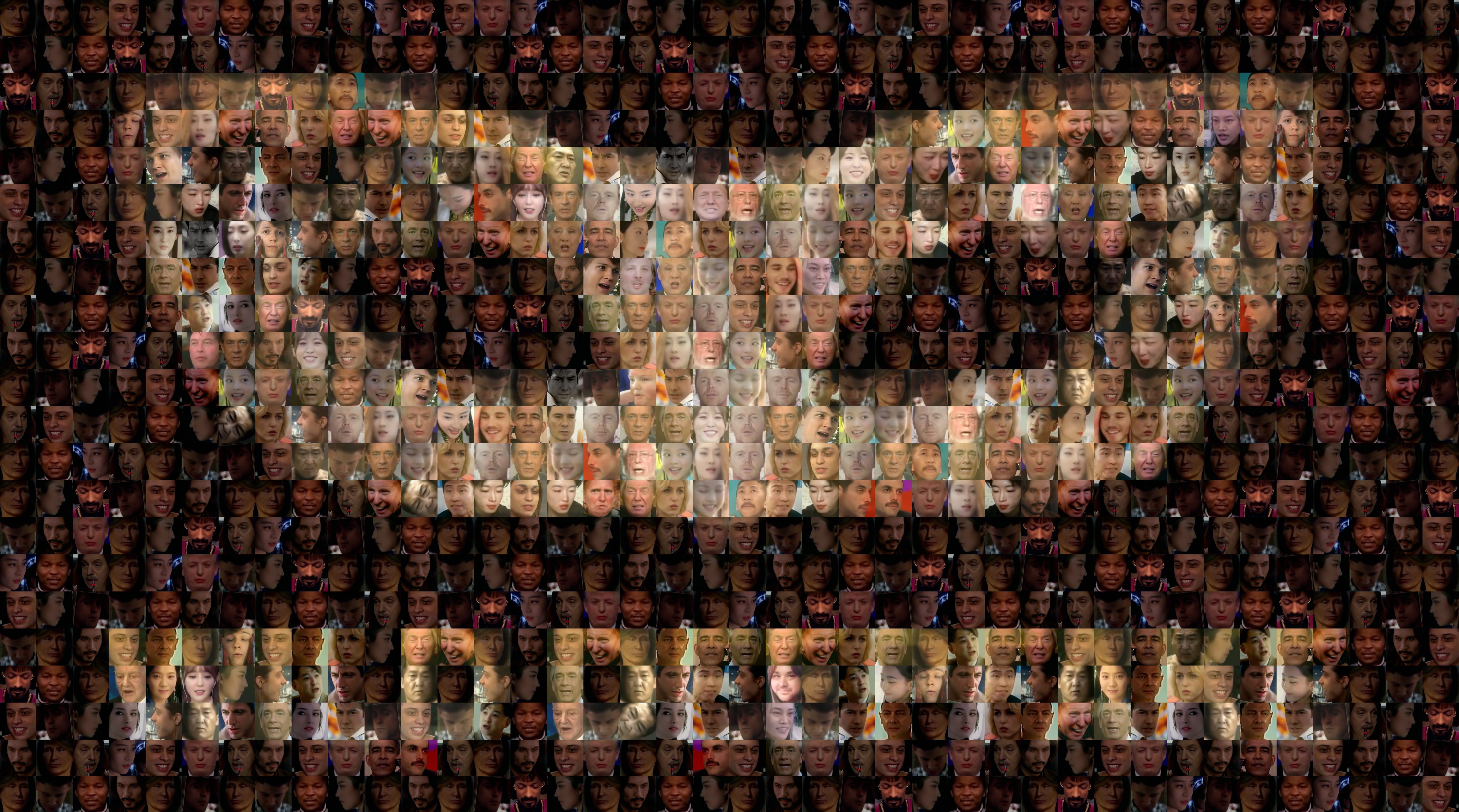}
  \captionof{figure}{WildDeepfake: a challenging real-word dataset for deepfake detection.}
  \label{fig:fake_mask}
\end{teaserfigure}

\maketitle

\section{Introduction}
    Deepfake (or "AI faceswap") refers to the set of deep learning-based facial forgery techniques that can swap one person's face in a video to another person. Face swap is not new, however, the emerging of deep learning techniques such as autoencoders and generative adversarial networks \cite{GAN} (GANs) makes face swap much easier and more convincing.
    Over the past few years, deepfakes have gone viral and a large number of deepfake videos (known as "deepfakes") have been crafted and uploaded to the internet. These fake videos have raised enormous public concerns for their huge risks to create political distress, blackmail someone or even fake terrorism events \cite{DFDC}. It is thus imperative to develop effective countermeasures to identify and reject deepfakes.
    
    One promising countermeasure against deepfakes is deepfake detection. However, training deepfake detectors generally requires a large amount of both real and deepfake videos. This drives the collection of several deepfake datasets such as Celeb-DF \cite{cela-DF}, UADFV \cite{UADFV}, Deepfake-TIMIT \cite{DF-TIMIT} and FaceForensics++ \cite{FaceForensics}. Recently, Google and JigSaw published a large dataset DeepfakeDetection \cite{DeepfakeDetection} in the 
   latest version of FaceForensics benchmark (eg. FaceForensics++) for deepfake detection. Another recent dataset was released by Facebook and Microsoft in the Deepfake Detection Challenge \cite{DFDC}. 
   Most of these datasets are collected following a similar process: 1) collecting source (real) videos, then 2) crafting deepfake videos (based on the source videos) using several popular deepfake softwares.
   Since the fake videos are crafted by the researchers rather than real-world deepfakes uploaded to the internet, we denote the fake videos in these datasets as \textbf{virtual deepfakes}. 
   Moreover, most of the source videos are filmed with a few volunteer actors in limited scenes. As such, virtual deepfakes may not fully represent the vast variety of wild deepfakes on the internet. We elaborate two potential weaknesses of existing virtual deepfake datasets as following:
    \begin{itemize}
    
        \item Lack of diversity. Virtual deepfakes contain limited types of scenes, only a few persons (most of the time, a single person) in each scene, and similar facial expressions/movements (mostly talking). In contrast, wild deepfakes can have more than 10 persons in one scene, and the scenes varies significantly cross different videos. Moreover, the deepfake techniques used to craft virtual deepfakes only cover the few popular ones. However, wild deepfakes are crafted by many different types, versions or even combinations of deepfake softwares. And most of the time, the exact software used to create a wild deepfake is unknown.
        
        \item Low quality. Via a preliminary inspection of virtual deepfakes, we find that many of the fake faces in these videos have obvious flaws. This may be because many virtual deepfakes are crafted in a short amount of time without careful adaptations for lighting, scene and a set of other factors. Consequently, the face regions in virtual deepfakes often have perceptible distortions such as jitters, blurs or strange artifacts. On the contrary, most wild deepfakes are deliberately tuned to have higher quality, may be via a long time of training on many high resolution face images.
    \end{itemize}

    Due to the above two weaknesses, detectors trained on virtual deepfake datasets may not fully generalize to wild deepfakes in the real world. To better support the development and evaluation of more effective deepfake detectors, in this paper, we introduce a new deepfake dataset that is collected completely from the internet: \textbf{WildDeepfake}. Figure 3e 
    illustrates several fake faces in our WildDeepfake dataset.
    In order to demonstrate the practical challenges in detecting wild deepfakes, we run extensive experiments with a set of baseline detection networks on both existing and our WildDeepfake datasets.
    We also propose two (eg. 2D and 3D versions) new Attention-based Deepfake Detection Networks (ADDNets) for more advanced deepfake detection.
    ADDNets exploit facial landmarks extracted by facial landmark detector to generate an attention mask to reweight the low-level features of a face, and then use reweighted low-level features to train either a 2D CNN detection network for image-level deepfake detection, or a 3D CNN detection network for sequence-level detection.
    In summary, our main contributions are:
    \begin{itemize}
        \item We collect and annotate a new challenging real-world dataset for deepfake detection: \textbf{WildDeepfake}. 
        Both the deepfake and real videos in WildDeepfake are collected purely from the internet. Compared to existing virtual deepfake datasets, WildDeepfake contains more diverse scenes, more persons in each scene and rich facial expressions.
    
        \item We conduct a systematic evaluation of a set of baseline detection networks on both existing and our WildDeepfake datasets, and show that these detectors all perform well on existing datasets yet poorly on WildDeepfake. This confirms that real-world deepfakes are indeed more challenging than virtual deepfakes.

        \item We propose two (eg. 2D and 3D) Attention-based Deepfake Detection Networks (ADDNets) against real-world deepfakes, and empirically verify the effectiveness of our ADDNets on both existing and the proposed WildDeepfake datasets.
    \end{itemize}

\begin{table*}[h]
\renewcommand\arraystretch{1.5}
\centering
 \caption{A summary of existing deepfake detection methods.}
\begin{tabular}{c|c|c|l}
\hline 
\textbf{Method}&\textbf{Dataset}&\textbf{Model}&\textbf{Claimed Performance}\\
\hline  \hline
MesoNet \cite{MesoNet}& Private web data&CNN&Detection rate: 98\%\\
\hline 
Guera~\etal \cite{Guera-Deepfake}&Private web data&\makecell[c]{CNN+LSTM}&Accuracy: 97.1\%\\
\hline 
FakeCatcher \cite{FakeCatcher}&\makecell[c]{FaceForensics++, \\ Private web data}&\makecell[c]{Traditional operator+CNN}&\makecell[l]{FaceForensics++ accuracy: 96\% \\ Private web data accuracy: 91.07\%}\\
\hline 
Li~\etal(1) \cite{ictu2018li}&Private web data&CNN+LSTM&Auc: 0.99\\
\hline
Li~\etal(2) \cite{li2018exposing}&\makecell[c]{UADFV,\\ Deepfake-TIMIT}&CNN&\makecell[l]{UADFV Auc: 0.974\\Deepfake-TIMIT(LQ) Auc: 0.999 \\ Deepfake-TIMIT(HQ) Auc: 0.932}\\
\hline
XceptionNet \cite{FaceForensics}&FaceForensics++&XceptionNet&\makecell[l]{Raw accuracy(Deepfake): 99.26\% \\HQ accuracy(Deepfake): 95.73\% \\LQ accuracy(Deepfake): 81.00\%}\\
\hline
Face X-ray \cite{X-ray}&\makecell[c]{Celeb-DF, \\DFDC preview, \\DeepfakeDetection \\and Faceforensics++}&\makecell[c]{FCN+Self-supervised learning}&\makecell[l]{FaceForensics++(Deepfake) Auc: 0.9917 \\DFDC preview Auc: 0.9540 \\DFDC Auc: 0.8092 \\Celab-DF Auc: 0.8058}\\
\hline
\hline
\end{tabular}
\end{table*}

\section{Related Work}
In this section, we briefly review several commonly used deepfake techniques and existing deepfake detection methods.

\begin{figure}[!htb]
    \centering
    \includegraphics[width=0.99\linewidth]{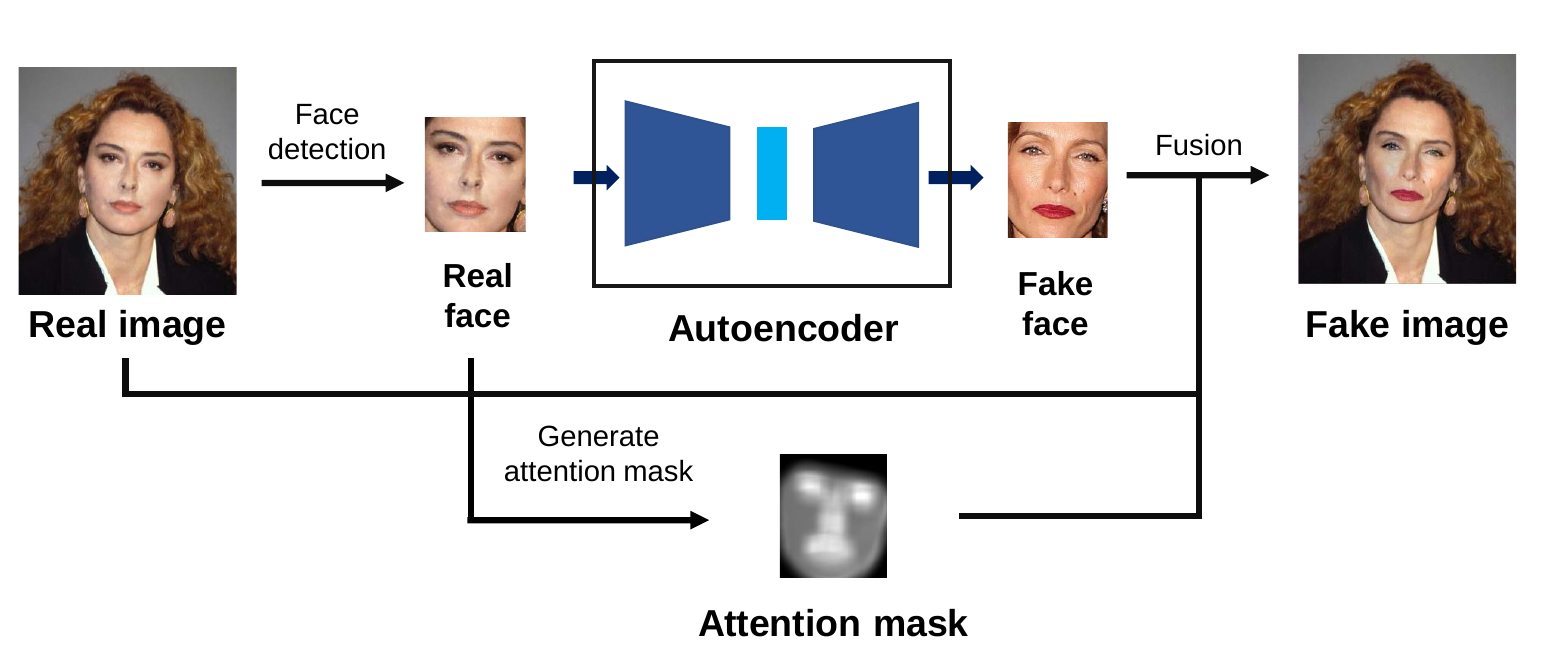}
       \caption{Illustration of the face swap process.}
    \label{fig:short}
    \end{figure}
    
\subsection{Deepfake Generation}
One commonly used deep learning technique for deepfake generation is the Generative Adversarial Networks (GANs) \cite{GAN}. There exist many open source deepfake softwares on GitHub, such as \emph{Faceswap-GAN} \cite{Deepfake-shaoanlu} and \emph{Faceswap} \cite{Deepfake}. Most of these deepfake softwares use an encoder-decoder architecture with one encoder and two decoders: the encoder learns the common features of the source (real) and the target (fake) faces, while the two decoders learn to generate the source and target faces separately. During the face swap process, the decoder associated with the source face takes the encoding of a target face and generate a fake source face. The attention mask of the source face is usually used to make the fake source face look more convincing via a fusion step. An overview of the face swap process is illustrated in Figure \ref{fig:short}. The generated fake faces can be further improved by using more high resolution face images (both source and target) to train both the encoder and decoders. Given a video, the face in each frame can be generated to replace the original face following the above face swap procedure.

\begin{table*}[!ht]
\renewcommand\arraystretch{1.15}
\centering
   \caption{A comparison of WildDeepfake with existing datasets for deepfake detection. LQ: low-quality; HQ: high-quality.}
   \label{tab:datasets}
\begin{tabular}{c|c|c|c|c|c}
\hline 
\textbf{Dataset}&\textbf{\makecell[c]{\#Real face\\sequences}}&\textbf{\makecell[c]{\#Fake face\\sequences}}&\textbf{\#Actors}&\textbf{\makecell[c]{Real video\\source}} &\textbf{\makecell[c]{Deepfake video\\source}}\\
\hline 
\hline

Deepfake-TIMIT&320&\makecell[c]{LQ:320\\HQ:320}&32&VidTIMIT Dataset \cite{VidTIMIT} & Manually crafted \\
\hline
\makecell[c]{FaceForensics++\\(Deepfake)}&\makecell[c]{Raw:1,000\\HQ:1,000\\LQ:1,000}&\makecell[c]{Raw:1,000\\HQ:1,000\\LQ:1,000}&977&YouTube & Manually crafted\\
\hline
Celab-DF v2&590&5,639&59&YouTube & Manually crafted\\
\hline
DeepfakeDetection&363&3,068&28&Volunteer Actors & Manually crafted\\
\hline
DFDC-preview&1,131&4,113&66&Volunteer Actors & Manually crafted\\
\hline
DFDC&$\sim$ 20,000& $\sim$100,000&-& Volunteer Actors & Manually crafted \\
\hline
\textbf{WildDeepfake(ours)}& \textbf{3,805} & \textbf{3,509} & \textbf{-} & \textbf{Internet} & \textbf{Internet} \\
\hline
\hline
\end{tabular}
\end{table*}

\begin{figure*}[!ht]

    \begin{tabular}{m{2.3cm} m{15.8cm}}
            \centering \begin{small}(a) Deepfake-TIMIT\end{small}&
            \includegraphics[width=0.84\textwidth]{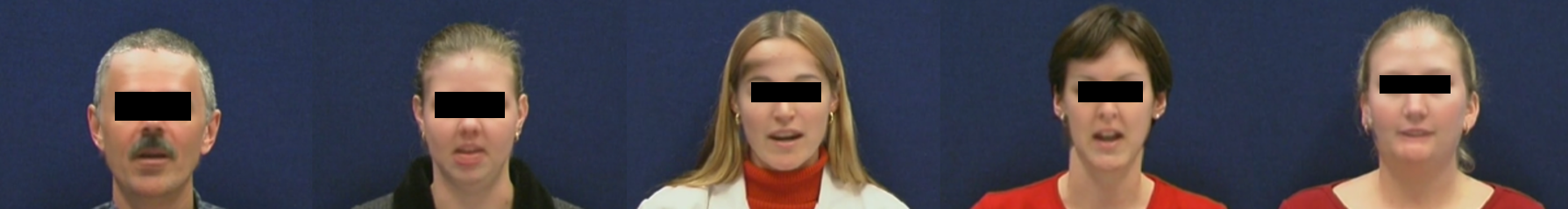}\\
            \centering \begin{small}(b) FaceForensics++\\(Deepfake)\end{small}&
            \includegraphics[width=0.84\textwidth]{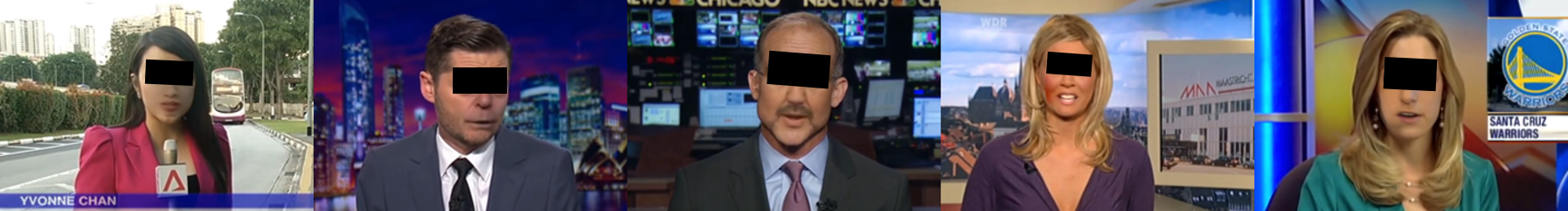}\\
            \centering \begin{small}(c) DFDC\end{small}&
            \includegraphics[width=0.84\textwidth]{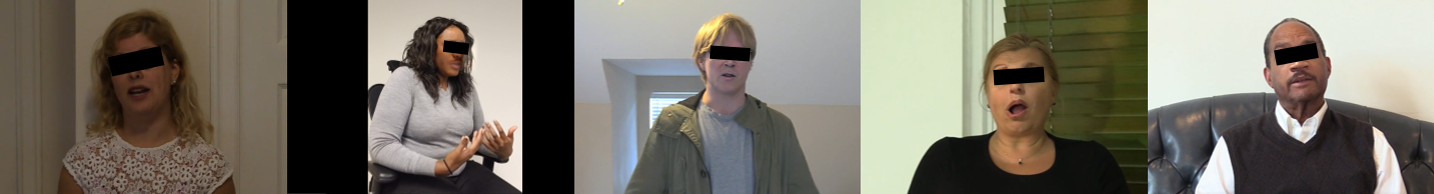}\\
            \centering \begin{small}(d) DeepfakeDetection\end{small}&
            \includegraphics[width=0.84\textwidth]{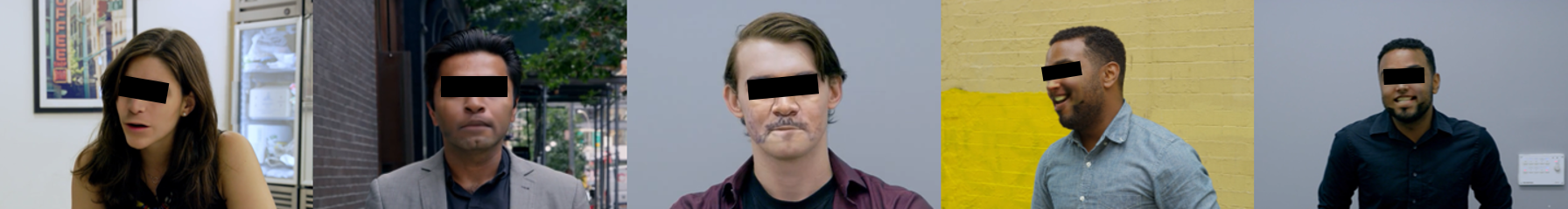}\\
            \centering \begin{small}(e) \textbf{WildDeepfake}\end{small}&
            \includegraphics[width=0.84\textwidth]{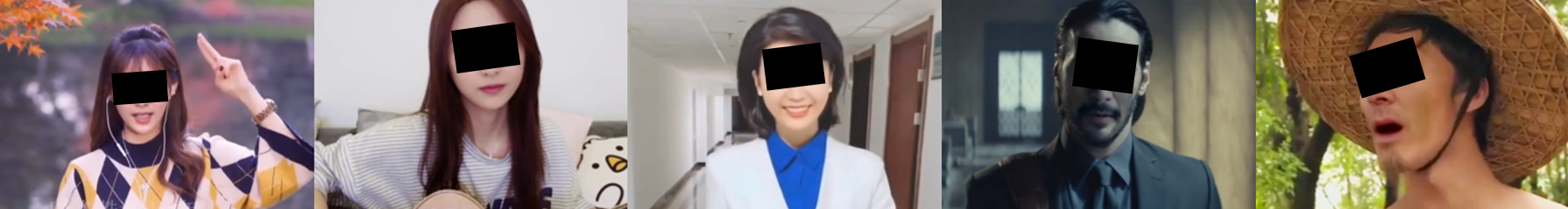}\\
    \end{tabular}
    \caption{WildDeepfake versus 5 existing datasets. There are more diverse scenes in WildDeepfake and the fake faces look more realistic, reflecting the challenging real-world scenario. To protect privacy, we block the eye regions of the fake images.}
\label{fig:example}
\end{figure*}

\subsection{Deepfake Detection}
A number of methods have been proposed to detect deepfake videos. Afchar~\etal proposed the MesoNet \cite{MesoNet}, which uses three shallow (a few number of layers) networks to examine the mesoscopic properties of face images. Güera~\etal \cite{Guera-Deepfake} revealed that the frame sequence of deepfake videos have unique characteristics, which differentiates them from unmodified videos. Therefore, they proposed to use CNN to extract features of video frames, then use an LSTM network to perform sequence prediction \cite{Guera-Deepfake}.
Ciftci~\etal proposed the FakeCatcher \cite{FakeCatcher} for deepfake detection. FakeCatcher exploits the difference of biological signals hidden in videos to distinguish fake videos from real videos. Li~\etal introduced two different methods \cite{li2018exposing,ictu2018li} to identify possible "artifacts" or eye blinking defects in deepfake videos. Motivated by the observation that XceptionNet \cite{XceptionNet} has better sensitivity to deepfake images, R{\"{o}}ssler~\etal used XceptionNet to detect deepfake images \cite{FaceForensics}. Recently, Li~\etal proposed the Face X-ray \cite{X-ray} to detect the trace of modification around the boundary regions of fake faces. 
Note that many of the above methods require pixel-level or image-level ground truth, which is not obtainable for real-world deepfakes.
In this paper, we will test those methods that do not rely on pixel/image-level annotations, i.e., XceptionNet and MesoNets. We conduct a systematic evaluation of XceptionNet, MesoNets and a set of conventional CNN networks on both existing and WildDeepfake datasets.

\section{Datasets for Deepfake Detection}
In this section, we first summarize existing deepfake datasets, then introduce the collection and annotation process of our WildDeepfake dataset.

\subsection{Existing Deepfake Datasets}
The UADFV \cite{UADFV} dataset contains 45 real videos and 45 deepfake videos, with the deepfake videos were crafted based on the real videos by applying some deepfake techniques. The Deepfake-TIMIT \cite{DF-TIMIT} dataset was created based on the VidTimit dataset: 320 low quality and 320 high quality deepfake videos were crafted based on 320 real videos from VidTimit. The FaceFornesics++ \cite{FaceForensics} dataset has 1,000 real videos collected from YouTube, based on which 1,000 deepfake videos were generated by applying each of the 4 face modification techniques: Deepfake \cite{Deepfake}, Face2Face \cite{Face2Face}, Faceswap \cite{FaceSwap} and Neural Texture \cite{NeuralTexture} (eg. overall 4,000 face modification videos were created). These fake videos produce 1.8 million manipulated face images. Recently, Google and JigSaw released the DeepfakeDetection \cite{DeepfakeDetection} dataset: 363 real videos were filmed with the assistance of 28 volunteer actors, based on which over 3,600 deepfake videos were then generated using a few deepfake techniques. More recently, AWS, Facebook, Microsoft, the Partnership on AI’s Media Integrity Steering Committee, and a number of academics collected and published a large-scale deepfake dataset for the Deepfake Detection Challenge (DFDC) \cite{DFDC}. DFDC dataset consists of $\sim$20,000 real videos filmed with hundreds of actors, and over 10,0000 fake videos generated using varies deepfake techniques. Table \ref{tab:datasets} summarizes these existing datasets. The deepfake videos in these datasets were all crafted by researchers applying a few popular deepfake techniques. They were not deliberately tuned to achieve the best visual effects, and some of the generated fake faces have obvious flaws. Detectors trained on these dataset may not generalize well to wild deepfakes. Next, we introduce our WildDeepfake dataset.

\begin{figure*}[t!]
\begin{minipage}{0.95\linewidth}
\centering
\small
\subfloat[Deepfake-TIMIT \protect\\ \protect\centering high-quality]{\label{fig:des_1}
   \includegraphics[width=.166\linewidth]{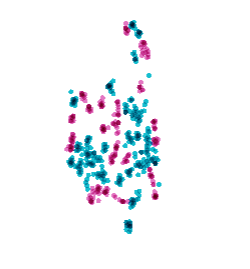}}
\subfloat[Deepfake-TIMIT \protect\\ \protect\centering low-quality]{\label{fig:des_2}
   \includegraphics[width=.166\linewidth]{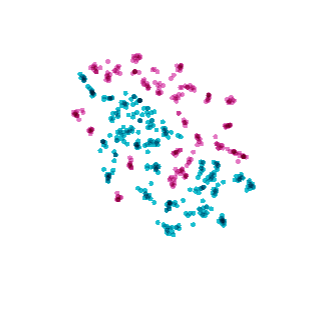}}
\subfloat[Faceforensics++ \protect\\ \protect\centering (Deepfake)]{\label{fig:des_3}
   \includegraphics[width=.166\linewidth]{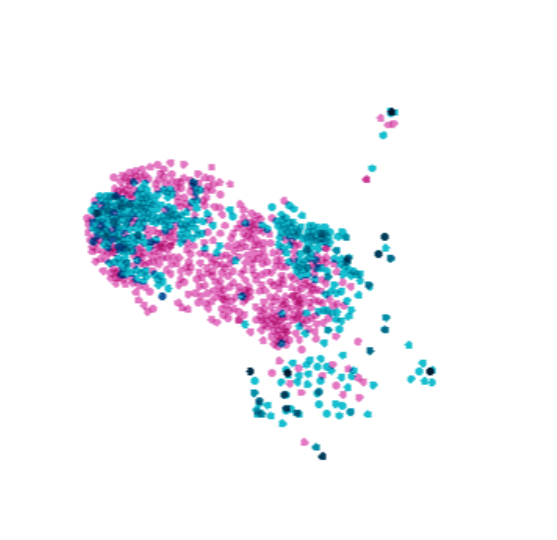}}
\subfloat[Deepfake Detection]{\label{fig:des_4}
   \includegraphics[width=.166\linewidth]{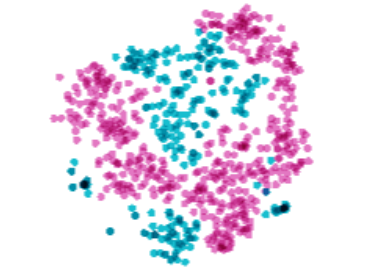}}
\subfloat[DFDC]{\label{fig:des_5}%
   \includegraphics[width=.166\linewidth]{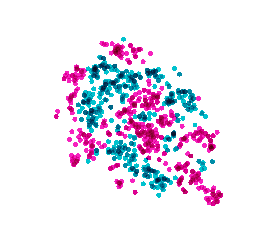}}
\subfloat[\textbf{WildDeepfake (ours)}]{\label{fig:des_6}
   \includegraphics[width=.166\linewidth]{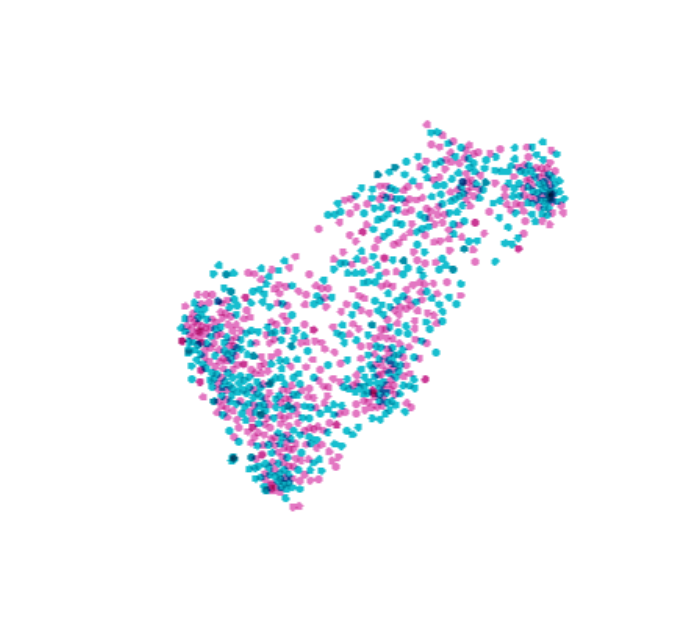}}
\caption{A feature perspective comparison of 6 deepfake datasets. We use an ImageNet-pretrained ResNetV2-101 network to extract features and t-SNE \cite{maaten2008visualizing} for dimensionality reduction.}
\label{fig:features}
\end{minipage}
\end{figure*}

\subsection{WildDeepfake Dataset}
We first collect over 1,200 deepfake videos from varies video-sharing websites. We collect these videos by searching their titles with keyword "deepfake". We remove those fake videos that were crafted using traditional face manipulations rather than deepfake techniques. We determine the type of the forgery technique by the title and description of the video. We then manually check and search the real video for each of the deepfake video. We remove those deepfake videos that do not have a real version. This leaves us \textbf{707} well-made deepfake videos from the internet.

\noindent\textbf{Data Processing.}
We use the Mtcnn \cite{Mtcnn} face detector to identify the face regions in each video frame. We then extract features for the face regions using an ImageNet-pretrained MobileNetV2 \cite{MobileNetV2} network. 
Next, we use the facial landmark extracted by dlib\cite{dlib} landmark detector to align all the faces in a face sequence. This avoids the negative impact of face orientation to the training of deepfake detectors.

\noindent\textbf{Face Sequence Annotation.}
We train 3 human annotators by explaining the background knowledge of deepfake generation, the common defects and characteristics of deepfake videos.
After training, the 3 annotators were asked to 1) label the type (eg. \emph{real}, \emph{fake} or \emph{unknown}) of each face sequence by checking the title of the collected video; 2) locate its source (real) video if a video is deepfake and check whether there are defects in the source video; 3) label the face sequence of the source video as \emph{real} if there are no obvious defects, otherwise label it as \emph{unknown}. 

We only save the face sequences of \emph{real} and \emph{fake} categories, while discard the \emph{unknown} categories.  We also discard those face sequences that have received different labels from the 3 annotators. Eventually, \textbf{1,180,099} face images of \textbf{7,314} face sequences from \textbf{707} videos were collected and annotated. It is worth mentioning that both the data processing and annotation is labour intensive and time consuming: labeling the 707 videos can take months.

\begin{figure*}
\centering
\subfloat[The proposed ADDNet-2D for image-level deepfake detection.]{\label{fig:image_level}%
\includegraphics[width=0.9\textwidth]{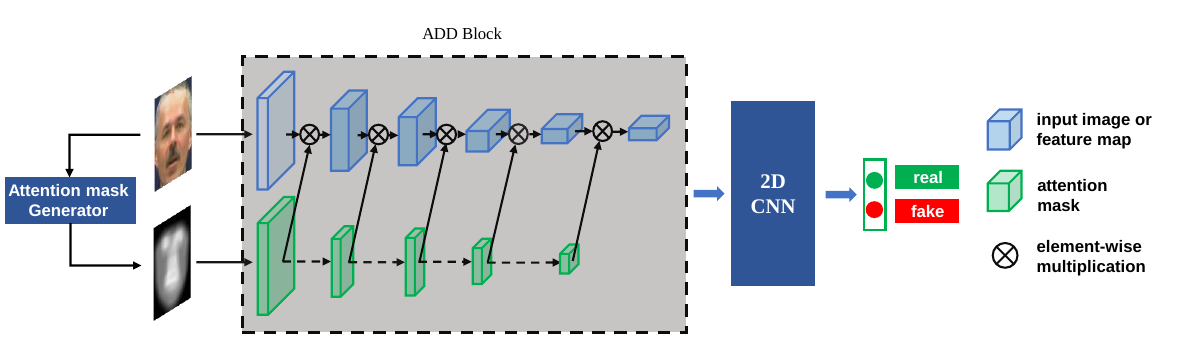}}\\
\subfloat[The proposed ADDNet-3D for sequence-level deepfake detection.]{\label{fig:face_sequence_level}
\includegraphics[width=0.78\textwidth]{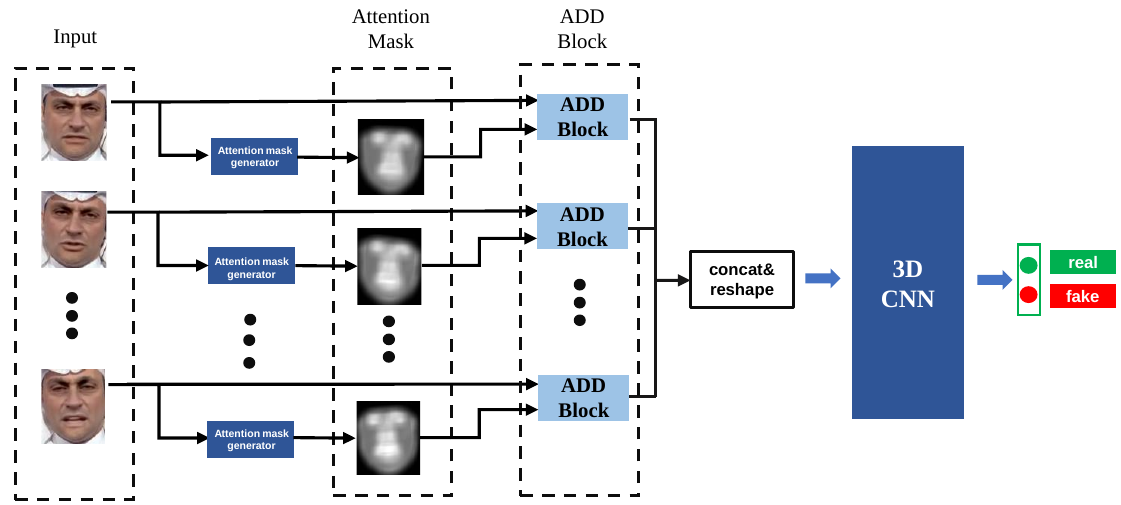}}
\caption{The structures of our ADDNet detection networks. The input size of 2D ADDNet is $W \times H \times C$, and that of the 3D ADDNet is $L \times W\times H \times C$: $W$: input width, $H$: input height, $C$: the number of channels, and $L$: sequence length.}
\label{fig:network}
\end{figure*}

The 7,314 face sequences in our dataset were further splitted into 6,508 for training and 806 for testing. This was done based on the similarities between the face sequences, which to some extent, ensures the training set having different face sequences from the test set. Figure \ref{fig:example} illustrates several deepfake video frames in varies datasets. Figure \ref{fig:features} shows the difference of the 6 datasets with respect to the features of the face images extracted by an ImageNet-pretrained ResNetV2-101 network. The major characteristics of our WildDeepfake dataset can be summarized as follows:

\begin{itemize}
    \item Both the real and deepfake videos in WildDeepfake were collected from the internet.
    \item The video contents in WildDeepfake are more diverse: a variety of activities (eg. broadcasting, movies, interviews, talks, and many others), diverse scenes, backgrounds and illumination conditions, and different compression rates, resolutions and formats.
    \item The deepfake videos in WildDeepfake were well-made, possibly owing to longer time of training with many high quality face images. 
\end{itemize}

\section{Proposed ADDNets for Deepfake Detection}
In this section, we introduce the proposed Attention-based Deepfake Detection Networks (ADDNets). We first formulate the detection problem as follows.

\subsection{Problem Formulation}
Given a deepfake dataset $\D = \{(\xx, y)^{(i)}\}_{i=1}^n$ with $\xx \in \X \subset \mathbb{R}^{F \times W \times H \times C}$ and $y \in \Y = \{0, 1\}$ denoting a video and its class label. $F$, $W$, $H$ and $C$ represent the number of frames, frame width, frame height and color channel, respectively. Deepfake datasets consist of two types of videos: real videos with class label $y=0$ and deepfake videos with class label $y=1$. The goal of deepfake detection is to train a binary classifier $f$ (as represented by a deep neural network) that maps the video space to the class space: $f: \X \rightarrow \Y$. This can be achieved by minimizing the classification error of $f$ on training data $\D$:
\begin{equation}\label{eq:objective}
    \argmin_{\theta} \E_{(\xx, y) \in \D} \ell(f(\xx), y),
\end{equation}
where $\ell$ is a loss function such as the commonly used Cross Entropy (CE) loss, and $\theta$ are the trainable parameters of network $f$. 

In this paper, we focus on training a CNN detection network $f$.
In practice, the input videos (both real and deepfake) are processed to extract face images, which can then be used to train a detection network. The reason why not directly train on raw video or video frames is that deepfakes only alter the face region.
Generally, there are two levels of deepfake detection networks: image-level and sequence-level. Image-level detection networks work on individual face images without considering the sequential information contained in the face sequence, while sequence-level detection networks work on the full face sequences.
Next, we propose two detection networks for image- and sequence-level deepfake detection.

\subsection{Proposed Detection Networks}\label{sec:addnet}
While existing works are mostly focused on identifying the flaws in a face image/sequence, most wild deepfake videos are carefully tuned to have no obvious flaws. 
Motivated the observation that many deepfake techniques use an attention mask to fuse the target (fake) face into the source (true) face (see Figure \ref{fig:short}), we propose to exploit the attention mask to better differentiate between the real and the fake faces. 
The attention-based face fusion operation can be defined as:
\begin{equation}
     O = t\odot(E-A) + g\odot A,
\end{equation}
where $\odot$ represents the element-wise multiplication, $g$ is the face generated by a neural network, $A \in [0,1]^{d}$ is the attention mask used in face fusion, $E$ is the identity matrix which has the same dimension $d$ as matrix $A$, $t$ is the source (true) face and $O$ is the output fake face.
Attention mask $A$ defines the key areas of the face, such as eye, nose, and mouth.

The proposed ADDNets are illustrated in Figure \ref{fig:network}.
We introduce two versions of ADDNet: a 2D ADDNet (ADDNet-2D) for image-level deepfake detection and a 3D ADDNet (ADDNet-3D) for sequence-level deepfake detection.

\subsubsection{ADDNet-2D}
As illustrated in Figure \ref{fig:image_level}, the 2D ADDNet consists of an ADD block, which is followed by a 2D CNN network and a classification layer. It takes one face image and the attention mask of the face image as inputs, and outputs the probabilities of the input face image being real (class 0) or fake (class 1). 
The input attention mask is generated via an Attention Mask Generation module, which will be explained in detail below.
Note that the attention mask generated here are different to the ones used to craft deepfakes, which are unknown.
The ADD block follows an XceptionNet architecture, which learns different levels of features of the the face image. Different to conventional CNN network, the features at the intermediate layers of the ADD block are adjusted by the scaled attention masks (in green color). More specifically, we scale the input attention mask to match the output resolution (eg. width and height) of a particular layer using average pooling, then apply an element-wise multiplication between the scaled attention mask and the feature map of that layer.
This allows the use of attention to adjust the feature map at different abstraction levels. We find that this is generally more effective than only applying the attention mask to the input layer. 
The output of the ADD block is then passed into a conventional 2D CNN network for classification. The output layer of the entire network is convolutionally (not fully) connected to the second last layer, and has two neurons corresponding to the two classes (eg. real vs fake). 

\subsubsection{ADDNet-3D}
We also propose a 3D ADDNet for sequence-level detection. As illustrated in Figure \ref{fig:face_sequence_level}, the network has multiple ADD blocks, the outputs of which are concatenated and reshaped before passing into a 3D CNN network for classification. Note that all ADD blocks in ADDNet-3D share the same weights. The network takes inputs of face sequences and its corresponding attention mask sequences. We denote the sequence length as $L$. For each face image in the sequence, we generate its attention mask using the same attention mask generation module as used in ADDNet-2D. Each pair of a face image and its attention mask is processed by one separate ADD block, which also has the same structure as is used in ADDNet-2D.

\begin{figure}[!htb]
\vspace{-0.1 in}
\centering
\includegraphics[width=0.98\linewidth]{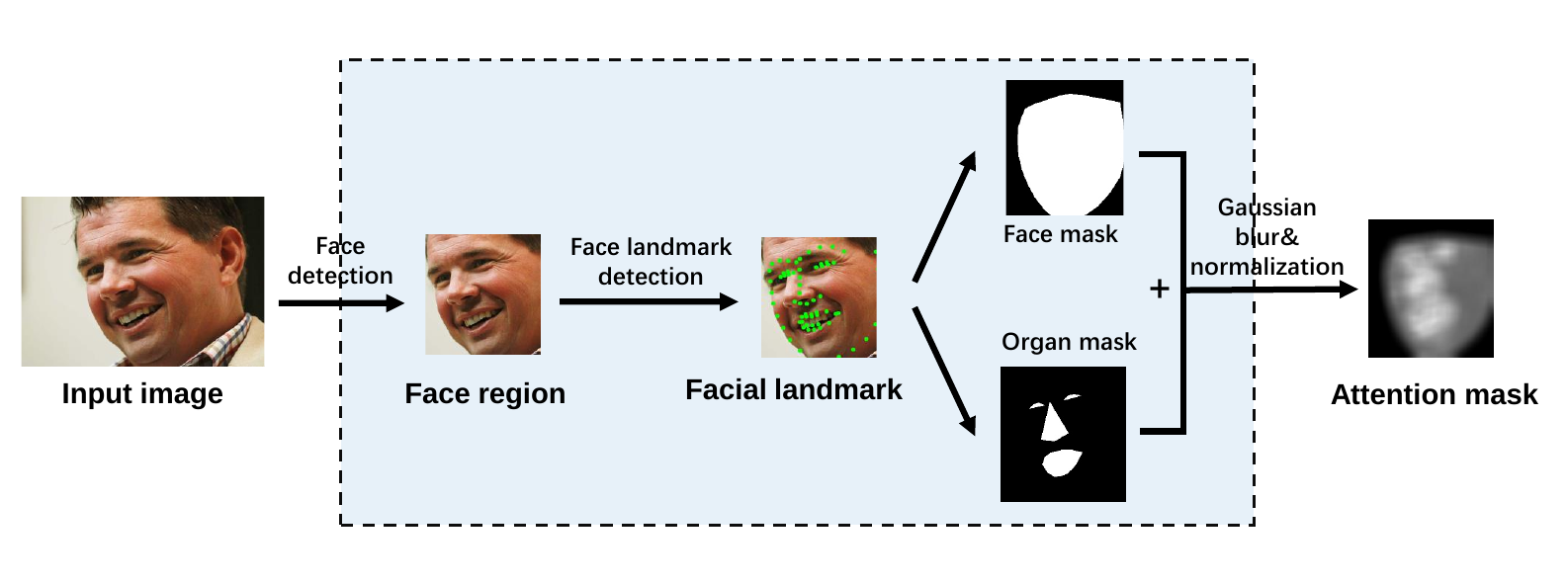}
\caption{The Attention Mask Generation module.}
\label{fig:mask_generation}
\vspace{-0.1 in}
\end{figure}

\subsubsection{Attention Mask Generation Module}
The attention mask generation process is shown in Figure \ref{fig:mask_generation}. 

Given a face image $x$, the attention mask is generated in 4 steps. We first apply a landmark detection method to identify the 68-points facial landmark of the face area and align the face according to the landmark. We then use the landmark to generate a binary \emph{face mask} that contains the entire face region. We apply the same process to generate a second \emph{organ mask} that contains the eyes, nose and mouth. We use Gaussian blur to smooth the edges of both masks (eg. face and organ masks). Finally, we add up the two masks and normalize it into the value range of $[0,1]$. The generated attention mask is used in the ADD block to adjust the feature maps of the face image, as we have introduced above.

Note that sequence-level detection can also be achieved by an ADDNet+LSTM architecture, which can be obtained by replacing the 3D CNN network in 3D ADDNet (see Figure \ref{fig:face_sequence_level}) by an LSTM network. The main difference between our ADDNets and existing detection networks is the application of attention-based feature adjustments at multiple layers of the ADD block.

\begin{table*}
\begin{center}
   \caption{Image-level detection accuracy of different 2D detection networks. LQ: low quality, HQ: high quality.}
   \label{image_acc}
\begin{tabular}{c|c|c|c|c|c|c}
\hline 
\textbf{Network}&\textbf{DFD} &\textbf{\makecell[c]{DF-TIMIT \\LQ}} &\textbf{ \makecell[c]{DF-TIMIT \\ HQ}}&\textbf{\makecell[c]{FF++(Deepfake) \\LQ}}&\textbf{\makecell[c]{FF++(Deepfake) \\HQ}}&\textbf{\makecell[c]{Wild-\\Deepfake}}\\
\hline  \hline
AlexNet \cite{AlexNet}&84.37\%&94.77\%&83.22\%&90.58\%&95.52\%&60.37\%\\
\hline 
VGG16 \cite{vgg}&90.02\%&98.73\%&76.92\%&90.19\%&98.89\%&60.92\%\\
\hline 
ResNetV2-50 \cite{ResNetv2} &83.68\%&94.88\%&89.51\%&\textbf{90.91\%}&98.59\%&63.99\%\\
\hline
ResNetV2-101 \cite{ResNetv2}&81.77\%&94.78\%&87.09\%&88.67\%&98.72\%&58.73\%\\
\hline
ResNetV2-152 \cite{ResNetv2}&83.15\%&95.68\%&88.27\%&88.00\%&97.57\%&59.33\%\\
\hline 

Inception-v2 \cite{inceptionv2}&72.64\%&90.30\%&77.92\%&89.44\%&96.67\%&62.12\%\\
\hline 
MesoNet-1 \cite{MesoNet}&75.95\%&92.07\%&79.98\%&81.97\%&96.40\%&60.51\%\\
\hline
MesoNet-4 \cite{MesoNet}&85.02\%&91.18\%&83.71\%&87.75\%&97.04\%&64.47\%\\
\hline 
MesoNet-inception \cite{MesoNet}&70.71\%&97.85\%&85.28\%&84.82\%&97.16\%&66.03\%\\
\hline
XceptionNet \cite{XceptionNet}&85.82\%&\textbf{99.65\%}&\textbf{99.91\%}&90.25\%&99.62\%&69.25\%\\
\hline 
\makecell[c]{\textbf{ADDNet-2D}\\\textbf{(ours)}}&\textbf{97.51\%}&99.54\%&99.22\%&90.42\%&\textbf{99.82\%}&\textbf{76.25\%}\\
\hline 
\hline
\end{tabular}
\end{center}
\end{table*}

\begin{table*}
\begin{center}
   \caption{ Sequence-level detection accuracy of different 3D detection networks.  LQ: low quality, HQ: high quality.}
   \label{seq_acc}
\begin{tabular}{c|c|c|c|c|c|c}
\hline 
\textbf{Network} &\textbf{DFD} &\textbf{\makecell[c]{DF-TIMIT\\LQ}}&\textbf{
\makecell[c]{DF-TIMIT \\HQ}}&\textbf{\makecell[c]{FF++(Deepfake) \\ LQ}}& \textbf{\makecell[c]{FF++(Deepfake) \\HQ}}& \textbf{\makecell[c]{Wild-\\Deepfake}}\\
\hline \hline
P3D \cite{p3d}&70.16\%&76.71\%&62.25\%&67.05\%&75.23\%&53.20\%\\
\hline
C3D \cite{c3d}&73.18\%&94.44\%&82.38\%&87.72\%&95.00\%&55.87\%\\
\hline 
I3D \cite{i3d}&67.83\%&\textbf{96.38\%}&\textbf{89.85\%}&\textbf{93.18\%}&96.70\%&62.69\%\\
\hline 
\makecell[c]{\textbf{ ADDNet-3D}\\\textbf{(ours)}}&\textbf{94.93\%}&90.17\%&85.75\%&90.11\%&\textbf{98.30\%}&\textbf{65.50\%}\\
\hline
\hline
\end{tabular}
\end{center}

\end{table*}

\section{Experiments}\label{sec:experiments}
In this section, we provide a systematic evaluation of a set of detection networks on both existing deepfake datasets and our WildDeepfake.

\subsection{Experimental Settings}
\noindent\textbf{Datasets.} We consider three existing datasets DeepfakeDetection (DFD) \cite{DeepfakeDetection}, Deepfake-TIMIT (DF-TIMIT) \cite{DF-TIMIT} and FaceForensics++ (FF++) \cite{FaceForensics}. For DF-TIMIT and FF++ datasets, we consider both their low  quality (resolution) (LQ) and high quality (resolution) (HQ) versions. For FF++, we only consider its deepfake subset. We also test the detection networks on our WildDeepfake dataset. Overall, we run experiments on 6 datasets: DFD, DF-TIMIT LQ, DF-TIMIT HQ, FF++ LQ, FF++ HQ and WildDeepfake.

\noindent\textbf{Baseline Detection Networks.}
For image-level detection, we compare our ADDNet-2D with 10 detection networks including AlexNet, VGG16, ResNetV2-50/101/152, Inception-v2, XceptionNet \cite{XceptionNet}, MesoNet-1, MesoNet-4 and MesoNet-Inception. The first 6 networks are state-of-the-art CNN networks proposed for image classification. We consider these networks to test the detection performance if directly applying a conventional CNN network in deepfake detection. The XceptionNet, MesoNet-1 \cite{MesoNet}, MesoNet-4 and MesoNet-Inception are previously proposed detection networks for deepfake detection. For sequence-level detection, we compare our ADDNet-3D with P3D \cite{p3d}, C3D \cite{c3d} and I3D \cite{i3d}, which are three state-of-the-art 3D networks for video recognition. 
Note that, we did not consider those existing methods that require pixel-level ground truth or the whole image, which is not obtainable for wild deepfakes.

\noindent\textbf{Training Setting.}
 For image-level detection, we set the input size to $224 \times 224$. For sequence-level detection, we train and test with clips (one clip contains 50 frames, i.e., sequence length $L=50$) and set the input image size to $112 \times 112$. 
 All networks including both image- and sequence-level detection networks are trained using cross entropy loss and Adam optimizer \cite{Adam} with batch size 32. We set the initial learning rate to 0.0001 which is decayed by a factor of 0.9 after every 3K steps of training. All networks are trained for 40,000 iterations. We choose the model with best accuracy as the final model.

\subsection{Results and Analysis}
We take the detection accuracy of deepfake videos as a performance metric, and discuss the detection performance separately for image-level and sequence-level detection networks.

\subsubsection{Image-level Deepfake Detection}
Table \ref{image_acc} reports the detection accuracies of all 2D detection networks.
As can be observed, although with certain variations, the baseline networks all demonstrate a good performance on existing datasets, especially on the high-quality ones. Particularly, the best baseline network achieves a high accuracy of 90.02\% on DFD (by VGG16), 99.65\% on DF-TIMIT LQ (by XceptionNet), 99.91\% on DF-TIMIT HQ (by XceptionNet), 90.91\% on FF++ LQ (by ResNetV2-50) and 99.62\% on FF++ HQ (by XceptionNet). Among the baseline networks, the XceptionNet achieves the best performance on three out of the five existing datasets. The performance of these networks decreases drastically on WildDeepfake dataset: no baseline networks can achieve accuracy above $70\%$. This not only confirms that real-world deepfakes are indeed more difficult to be detected than virtual deepfakes, the effectiveness of detectors developed on virtual deepfake datasets can be limited when applied to detect wild deepfakes.

Our proposed ADDNet-2D achieves a comparable performance to the best baseline networks on existing datasets, and a significantly better performance on the more challenging WildDeepfake dataset. Particularly, on DFD dataset, ADDNet-2D outperform the best network XceptionNet by 11\%, while on other existing datasets it exhibits an accuracy that is within 1\% if not better than the best baseline networks. On WildDeepfake dataset, ADDNet-2D outperforms the best baseline network XceptionNet by 7\%. The consistent and superior performance of our ADDNet-2D network verifies the importance of attention-based feature adjustment for deepfake detection. Note that, even our ADDNet-2D can not fully address the challenge of detecting wild deepfakes: the detection accuracy is only 76.25\%. 

\subsubsection{Sequence-level Deepfake Detection}
The detection accuracies of different 3D detection networks for sequence-level detection are reported in Table \ref{seq_acc}. Among the baseline networks, I3D achieves the best performance on almost all tested datasets, except DFD where C3D is more effective. Particularly, I3D achieves a high detection accuracy of $> 89\%$ on the four DF-TIMIT and FF++ datasets, while C3D has an accuracy of 73.18\% on DFD. Again, they all drop significantly on WildDeepfake with a much lower accuracy of $< 63\%$.
Compared to the I3D and C3D, P3D has the worst performance, which indicates that the pseudo 3D convolutions used in P3D are not sensitive enough to small deepfake modifications.

For our ADDNet-3D, it is not as effective as I3D on two DF-TIMIT datasets and the low quality FF++, although it achieves a surprisingly higher accuracy of 94.93\% on DFD and outperforms I3D on higher quality FF++.
On WildDeepfake, ADDNet-3D demonstrates a 65.50\% accuracy, which is $\sim 2\%$ higher than the best baseline model I3D.
However, this performance is much lower than ADDNet-2D (76.25\%).  Compared to 2D detection networks, we find that 3D networks are generally less effective. 
One possible reason for the performance degradation of 3D detection networks is that the temporal information contained in deepfake face sequences are also distorted by the frame-by-frame generation of the fake faces. And such variations are very likely inconsistent across different frames or videos. This indicates that the temporal information in deepfake videos should be treated differently to that in real videos to improve the accuracy of sequence-level deepfake detection.

\section{Conclusion}
In this paper, we proposed a challenging real-world dataset \textbf{WildDeepfake} for deepfake detection. WildDeepfake dataset consists of \textbf{1,180,099} images of \textbf{7,314} face sequences extracted from \textbf{707} videos (both deepfake and real). Compared with existing virtual deepfake datasets, our WildDeepfake dataset was collected completely from the web, thus contains more diverse scenes, faces and activities. Moreover, the deepfake videos in our dataset are of high quality. WildDeepfake can serve as a useful supplementary to existing datasets to support the development and evaluation of more effective deepfake detectors against real-world deepfakes. We conducted a systemic evaluation of a set of baseline networks on both existing and our WildDeepfake datasets, and found that WildDeepfake is indeed a more challenging dataset where the performance of baseline detectors can decrease drastically.
We also propose two Attention-based Deepfake Detection Networks (ADDNets) to leverage the attention-based feature adjustment for more accurate deepfake detection. Our 2D version of ADDNet (eg. ADDNet-2D) demonstrated a better or at least a comparable performance to the state-of-the-art, consistently across all existing and our WildDeepfake datasets. We believe that, with our WildDeepfake dataset and ADDNets, more advanced countermeasures against real-world deepfakes can be developed in the future.

\newpage

\typeout{get arXiv to do 4 passes: Label(s) may have changed. Rerun}

\bibliographystyle{ACM-Reference-Format}
\balance
\bibliography{samples}

\end{document}